# Supervised Vector Quantized Variational Autoencoder for Learning Interpretable Global Representations*

Yifan Xue, Michael Ding and Xinghua Lu†

*Department of Biomedical Informatics, University of Pittsburgh*
*Pittsburgh, PA, 15206, United States*
†*Email: xinghua@pitt.edu*

Learning interpretable representations of data remains a central challenge in deep learning. When training a deep generative model, the observed data are often associated with certain categorical labels, and, in parallel with learning to regenerate data and simulate new data, learning an interpretable representation of each class of data is also a process of acquiring knowledge. Here, we present a novel generative model, referred to as the Supervised Vector Quantized Variational AutoEncoder (S-VQ-VAE), which combines the power of supervised and unsupervised learning to obtain a unique, interpretable global representation for each class of data. Compared with conventional generative models, our model has three key advantages: first, it is an integrative model that can simultaneously learn a feature representation for individual data point and a global representation for each class of data; second, the learning of global representations with embedding codes is guided by supervised information, which clearly defines the interpretation of each code; and third, the global representations capture crucial characteristics of different classes, which reveal similarity and differences of statistical structures underlying different groups of data. We evaluated the utility of S-VQ-VAE on a machine learning benchmark dataset, the MNIST dataset, and on gene expression data from the Library of Integrated Network-Based Cellular Signatures (LINCS). We proved that S-VQ-VAE was able to learn the global genetic characteristics of samples perturbed by the same class of perturbagen (PCL), and further revealed the mechanism correlations between PCLs. Such knowledge is crucial for promoting new drug development for complex diseases like cancer.

*Keywords:* Supervised vector quantized variational autoencoder; Generative model; Supervised deep learning; LINCS.

## 1. Introduction

Deep generative models have achieved unprecedented success in many computer application areas including computer vision, natural language processing, and bioinformatics. Their blooming is attributed to the capability of mimicking the data-generating process and learning latent representations that capture the statistical characteristics of individual training data. When the task is to learn global representations for major statistical structures in subgroups of data, current approaches usually consists of two steps: first learning a latent representation for each individual observation using autoencoder-based models [1-6]; and second identifying clusters in the newly-learned feature space and treating the centroid of each cluster as the representation for the subgroup of data. Since both steps are completed via unsupervised approaches, this comes with two major problems: first, it can be hard to know what information the latent representations capture; and second, the resulting clusters may be inconsistent with previously known intrinsic divisions of the data, both giving rise to less interpretable global representations.

* This work is supported by the National Human Genome Research Institute, Grant No. U54HG008540 via the trans-NIH Big Data to Knowledge (BD2K) Initiative (http://www.bd2k.nih.gov), by the National Library of Medicine, Grant No. R01LM012011, and by the Pennsylvania Department of Health, Grant No. 4100070287.

Models that can reconstruct the data generation process and learn interpretable global representations for subgroups of data that share common statistical structures are of particular interest in the domain of systems biology. Particularly, if a model can learn to accurately regenerate biological data produced under different cellular perturbations, e.g., gene expression or proteomics of cells under different stresses, the model must have learned a representation of the cellular signaling system responding to the perturbation. Such representations shed light on the mechanisms that distinct perturbations take in perturbing different cellular processes, which can be used to promote drug development for complex genetic diseases like cancer.

In this work, we introduce a novel generative model, the Supervised Vector-Quantized AutoEncoder (S-VQ-VAE), which integrates the two sequential steps for learning global representations into one framework by adding a supervised learning component to the standard Vector Quantized AutoEncoder (VQ-VAE) [7]. In particular, we extend VQ-VAE by utilizing the vector quantization (VQ) technique to discretize the latent variable space into multiple mutually exclusive subspaces and projecting data from each class into its own assigned subspace. In our model, the class label of each training data point is used to supervise the projection step, mapping the individual latent representation derived by the unsupervised encoder component to the embedding code that represents the centroid of the subspace. The embedding code of a class and the latent representations of input data are updated simultaneously so that the code learns to capture the primary characteristics of the class. Since the role of each embedding code is defined in advance, the code can be well interpreted after the training converges. The learned global representations can be used for both standard deep learning tasks such as data classification and generative model specific tasks such as data generation. Furthermore, they can be used to reveal the associations between different classes. These novel capabilities are previously not achievable by either conventional supervised or unsupervised generative models. We applied the S-VQ-VAE model on gene expression data from the Library of Integrated Network-Based Cellular Signatures (LINCS) [8, 9]. As we will show below, the global representation learned for each group of samples perturbed by the same class of perturbagen (PCL) provides knowledge about the mechanisms of action shared between PCLs.

## 2. Related Works

As indicated by its name, S-VQ-VAE is extended from the VQ-VAE model by leveraging the power of supervised learning. The VQ-VAE model was proposed by Oord et al. in 2017 [7] as a new way of training Variational AntoEncoders (VAEs) [5, 6] with discrete latent variables; the VAE model is itself a variation of the ordinary autoencoder. Although autoencoders have been widely used for learning reduced dimension representations for input data, they are generally considered less successful as a "generative" model, for the posterior distribution $q(z|x)$ of the latent variables $z$ given the input data $x$ is often intractable. VAEs exploit the idea of variational inference to approximate the posterior distribution of the latent variables with a variational distribution *p(z)*, thus facilitating the generation of new data. From VAEs, a continuous latent representation vector of *p(z)* is learned for each data point. For many modalities in nature, however, like the words in a sentence and the symbols in speech, a discrete representation is usually more rational. Therefore, VQ-VAEs extend VAEs and learn discrete latent representations by discretizing the latent variable space using the VQ technique. In VQ-VAE, an embedding space $E \in R^{K \times D}$ composed of *K* embedding codes of length *D* is constructed, and an input data point is encoded and represented by multiple discrete codes by mapping its latent representation to the embedding space through nearest-neighbor look-up. Consequently, the distribution of the latent variables is discrete and learned in VQ-VAE rather than pre-defined as a variational distribution as in VAE. Eq. (1) gives the objective function for training a VQ-VAE. The first term in the function is the reconstruction loss (e.g., Mean Squared Error (MSE)) between the input data *x* and the reconstructed data. The reconstructed data is computed from the embedding code $e_k$ that is nearest to the latent representation $z_e(x)$ of the input data according to Eq. (2). The second term is the VQ objective, which computes a $l_2$ error between the $z_e(x)$ and $e_k$ thus to update the embedding code to get close to the latent representation. The third term, called the commitment loss, is to constrain the volume of the embedding space, forcing the latent representation commits to an embedding code rather than growing arbitrarily. The *sg* is the stop gradient operator, which is defined as identity during forward computation and has zero partial derivatives. With this operator, the operand will not be updated with the gradient from the local computation in which it participates. As a result, the VQ objective and the commitment loss together update $z_e(x)$ and $e_k$ in a bidirectional manner through training.

$$L = l_r(x, d(e_k)) + \|sg[z_e(x)] - e_k\|_2^2 + \beta\|z_e(x) - sg[e_k]\|_2^2 \quad (1)$$

$$k = argmin_j \|z_e(x) - e_j\| \quad (2)$$

The iterative update of the embedding codes through nearest-neighbor mapping in VQ-VAEs share many similarities with the k-means clustering algorithm [10]. In fact, VQ-VAEs suffers from the same problems as k-means in the arbitrary choice of *k* and the obscurity in understanding the distance between centroids of clusters. People often choose a large number of codes, e.g., $2_{12\text{-}16}$, to allow the codes to fully explore the latent space [10, 11]. This large embedding space slows down the training process, and result in codes containing redundant information.

Variations of generative models that combine supervised learning with unsupervised learning have also been proposed in recent years [12, 13]. Most of these models, however, requires additional auxiliary subnetworks to incorporate the class label information, and no explicitly global representation is learned for a class of input data as a whole.

## 3. S-VQ-VAE

In this work, we introduce the S-VQ-VAE model, which is extended from VQ-VAE by incorporating the label information of the input data for training the embedding space of VQ-VAEs in a supervised manner. Like the standard VQ-VAE, a S-VQ-VAE is composed of three parts, an encoder to generate the latent representation $z_e(x)$ given an input vector *x*, an embedding space to look up the discrete representation $z_q(x)$ based on $z_e(x)$, and a decoder to reconstruct the input data from $z_q(x)$. Suppose the encoder encodes the input data to an embedding code of length $D$, the embedding space $E$ is defined as $E \in R^{Y \times D}$, where $Y$ is the number of different classes of the input data. Each of the $Y$ embedding codes is designated to learn a global representation of one of the classes. Figure 1 shows the structure of S-VQ-VAE and the forward computational flow. In forward computation, an input $x$ is first converted to its latent representation $z_e(x)$. During training, $z_e(x)$ is replaced with $z_q(x)=e_y$ to pass to the decoder, where $e_y$ is the embedding code of the class $y$ of $x$. During testing, $z_e(x)$ is replaced by its nearest code $z_q(x)=e_k$ as given in Eq. (2). Note that we are not assuming a uniform distribution of the embedding codes as in the standard VQ-VAE [7]. Instead, the distribution of codes is determined by the input data with its discrete class labeling.

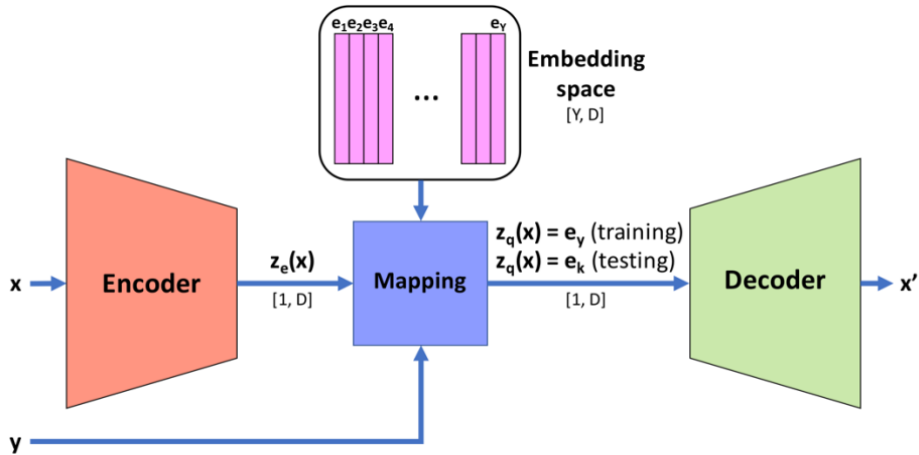

Fig. 1. The S-VQ-VAE model. The blue arrows mark the forward computational flow.

New data can be generated from S-VQ-VAE following two sampling steps, similar to ancestral sampling. First sample a target class $y$ from the distribution of class of the input data. This is to select the class that the new data should be generated from. Second, sample a latent vector $z \sim N(e_y, \sigma_2)$, where $\sigma_2$ is the covariance matrix of latent variables estimated from the input data of class $y$. A new data of class $y$ can then be generated by passing $z$ to the decoder of S-VQ-VAE. The generation process reflects another advantage of S-VQ-VAE compared to unsupervised generic models: we can determine what content the new data should present rather than interpret it afterward.

In order to achieve the goal of having a model that can learn individual representations through data reconstruction as well as learn a global representation for each class in a supervised manner, the objective function of S-VQ-VAE should contain a reconstruction loss to optimize the encoder and decoder, and a dictionary learning loss to update the embedding space. These make up the first two terms of the objective function as shown by Eq. (3). The form of reconstruction loss can be selected based on the data type, and here we used the MSE. Following the training protocol of standard VQ-VAEs [7], we chose VQ as the dictionary learning algorithm, which computes the $l_2$ error between $z_e(x)$ and $e_y$ thus updating the embedding code towards the latent representation of an input data of class $y$. To control the volume of the embedding space, we also added a commitment loss between $z_e(x)$ and $e_y$ to force the individual representation getting close to the corresponding global representation (third term in Eq. (3)).

$$L = l_r\left(x, d(e_y)\right) + \|sg[z_e(x)] - e_y\|_2^2 + \beta\|z_e(x) - sg[e_y]\|_2^2 - I(k \neq y)(\|sg[z_e(x)] - e_k\|_2^2 + \gamma\|z_e(x) - sg[e_k]\|_2^2) \quad (3)$$

In addition to making the latent representation and the corresponding global representation converge to each other, we added another two terms to force the latent representation of an input data to deviate from its nearest global representation $e_k$ if $k \neq y$ (i.e., misclassified through nearest-neighbor). As given in Eq. (3), the fourth term is another VQ objective which updates the embedding code of the mis-class. The last term, called the divergence loss, expands the volume of the embedding space in order to allow different classes to diverge from each other. Coefficients are applied to the commitment loss ($\beta$) and divergence loss ($\gamma$) to control the strength of regularization over the embedding space volume. According to pre-experiments using coefficients from [0, 1], the performance of the model is quite robust to these coefficients. For generating the results presented in this study, we used $\beta = 0.25$, and $\gamma = 0.1$. Note that the mapping step with either the class label or nearest neighbor has no gradient defined for it. As in training VQ-VAE, we approximate the gradient in a similar way as the straight-through estimator [14], by passing the gradient from the reconstruction loss from $z_q(x)$ directly to $z_e(x)$.

## 4. Experiments

We evaluated the utility of S-VQ-VAE on an image benchmark dataset, the MNIST dataset [15] and applied the model on gene expression data from the LINCS database [8, 9] for revealing associations between different types of perturbagen. The models for generating the results were implemented in Python using the *PyTorch* library [16], and illustrative Python codes for MNIST data are available from https://github.com/evasnow1992/S-VQ-VAE/.

The MNIST dataset we used was downloaded through the *torchvision* package in the *PyTorch* library for Python [16]. It contained 70,000 28 x 28 gray-scale images of handwritten digits from 0 to 9, from which 60,000 images were used as the training dataset and the remaining 10,000 were used as the test dataset. The encoder of our model for MNIST was composed of two convolutional layers of size {20, 50}, kernel size 5, and stride {2, 1} with ReLU activation function. The decoder had two transposed convolutional layers of size {50, 20}, kernel size {3, 5}, and stride {3, 1} with ReLU and Tanh activation functions, respectively. Since there were 10 classes of digits, the number of embedding codes was 10, and each code was a vector of length 3200 (8 x 8 convolved images x 50 channels). This architecture was selected based on pre-experiments using one to three convolution layers, 10 to 64 output channels on each layer, and with or without including max-pooling and/or dropout layers. The model was trained for 25 epochs, with batch size 256, and learning rate 1e-3. We also trained a standard VQ-VAE of the same architecture for comparison.

The LINCS project is a National Institutes of Health (NIH) Common Fund program that aims to create a network-based understanding of how human cells globally respond to various types of perturbations [8]. Specifically, each sample from a library of cell lines was treated with a single type of perturbation, like a small molecule or a single gene knockdown using RNAi or CRISPR, and different aspects of phenotype were measured, including mRNA expression profile. The expression profiling technique used in the LINCS project is a new approach known as the L1000 assay [9], were the "L1000" indicates the ~1,000 (978) landmark genes for which the expression levels are measured directly; the rest of the transcriptome is inferred from the landmark genes through a linear model. The LINCS database currently provides over one million expression profiles for samples from over a dozen cell lines perturbed by more than 10,000 different perturbations [9]. The expression data used for generating the results shown here are from the Gene Expression Omnibus (GEO), dataset GSE70138, which contains expression profiles of samples perturbed by small molecules. We extracted the level 5 data (moderated z-score of expression level) from one cell line MCF7 (breast cancer), to avoid mixing of cell line specific features, and we used the z-score (range [-10, 10]) of the 978 landmark genes as input to our model to avoid redundant correlations between the inferred genes. We only included samples where their perturbagen has been classified into one of the PCL by LINCS [9], enabling the use of PCL as the supervised label for each sample. We excluded samples perturbed by the proteasome inhibitor MG-132 as proteasome inhibitors generally change the expression profile in a more fundamental and inconsistent way compared to other perturbagens as we found out in pre-experiments. This final dataset contains 1,773 samples perturbed by small molecules from 75 PCLs. This dataset was further divided into a training dataset containing 9/10 of the data (1,596 samples) and a test dataset containing the rest (177 samples).

The encoder of the S-VQ-VAE we trained on the LINCS data contained a single hidden layer with 1,000 nodes and tangent activation function. The decoder was a reverse of the encoder with a tangent activation function on the output layer. The number of hidden layers and hidden nodes were selected based on pre-experiments with a wide range of architectures from one to two hidden layers and 200 to 1500 hidden nodes in each layer. The output data were rescaled from [-1, 1] to [-10, 10] before computing the reconstruction error. The embedding space contained 75 codes, one for each PCL. The model was trained for 1200 epochs, with batch size 128, and learning rate 1e-4. Again, a standard VQ-VAE with the same architecture was trained for performance comparison.

## 5. Results

### 5.1. *Utility evaluation on the MNIST data*

The performance of the S-VQ-VAE and the VQ-VAE with the same architecture on MINST data are shown in Table 1. We can see that with the same number of embedding codes (10), S-VQ-VAE achieved a lower reconstruction error for both the training and test datasets compared to the standard VQ-VAE. The perplexity of the embedding codes was close to 10 in S-VQ-VAE but only 1 in VQ-VAE, which indicates that all the 10 codes were used roughly uniformly in S-VQ-VAE while only one code was frequently used in VQ-VAE. In other words, without forcing each code to learn to represent a specific digit class, the standard VQ-VAE is much less efficient in exploring the expression power of the embedding space, and most data were mapped to a single code, and only this code was actively updated.

Table 1. Performance of S-VQ-VAEs and VQ-VAEs trained on MNIST and LINCS data.

| Model | Dataset | Perplexity | Training MSE | Test MSE |
|---|---|---|---|---|
| S-VQ-VAE | MNIST | 9.811 | 0.222 | 0.221 |
| VQ-VAE | MNIST | 1 | 0.272 | 0.272 |
| S-VQ-VAE | LINCS | 36.027 | 1.426 | 1.315 |
| VQ-VAE | LINCS | 40.263 | 1.236 | 1.553 |

To determine whether the embedding codes of S-VQ-VAE captured information of the digits that they were designated to learn, we visualized the codes by passing them through the decoder and generating images on the same scale as the input data. As expected, each decoded image takes the typical structure of a digit that the code learned to represent (Figure 2). The boundary of strokes is blurry compared to a real handwritten image, which reflects the part where people are most inconsistent in writing the digit. Note that these decoded images are different from global representation images generated by overlapping sample images from the same class. Each decoded image was generated from a unique feature vector that captures the crucial characteristics of a whole class.

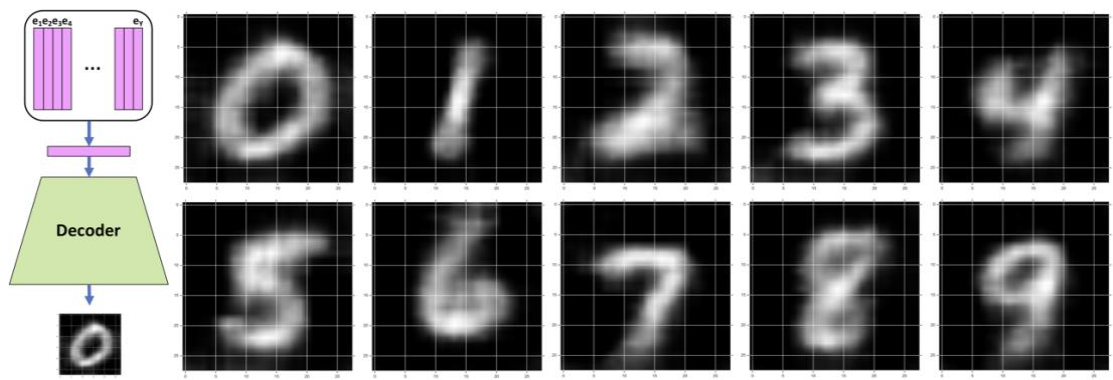

Fig. 2. The visualization of embedding codes for MNIST data. Each embedding code was trained in a supervised fashion to learn the global representation for one of the 10 digits. Left, the process for visualizing an embedding code. Right, the decoded images of the 10 embedding codes.

We then examined whether the global representations learned by S-VQ-VAE provide information about how different classes are correlated. This was done by examining the frequencies of nearest embedding codes for images of the same digit and computing the pairwise Euclidean distance between embedding codes. From Figure 3, it is clear that most images were correctly mapped to the code that was assigned to their classes. For the misclassified images, their nearest codes shed light on associations between different digits. For example, the second highest bar in the digit 4 subplot is the code for digit 9 (Figure 3). Correspondingly, the second highest bar in the digit 9 subplot is the code for digit 4. This suggests that the model found it a little difficult to distinguish between 4 and 9. Such associations are more clear on the heatmap (Figure 3 right), where the shortest distance is 4.970 between codes 4 and 9 as we expected, followed by 5.256 between codes 3 and 5, and 5.543 between codes 5 and 8. This indicates that in general, people write digit 4 in a more similar way to digit 9 relative to the other digits, and similarly for digits 3, 5 and 8. In Figure 4, we provide some example images for digit 4 and 9. Indeed, sometimes people write these two digits in the same stroke order with almost indistinguishable shapes. The embedding codes learned with S-VQ-VAE correctly captured these intrinsic similarities in human handwriting, and reflected them in the distances between code vectors.

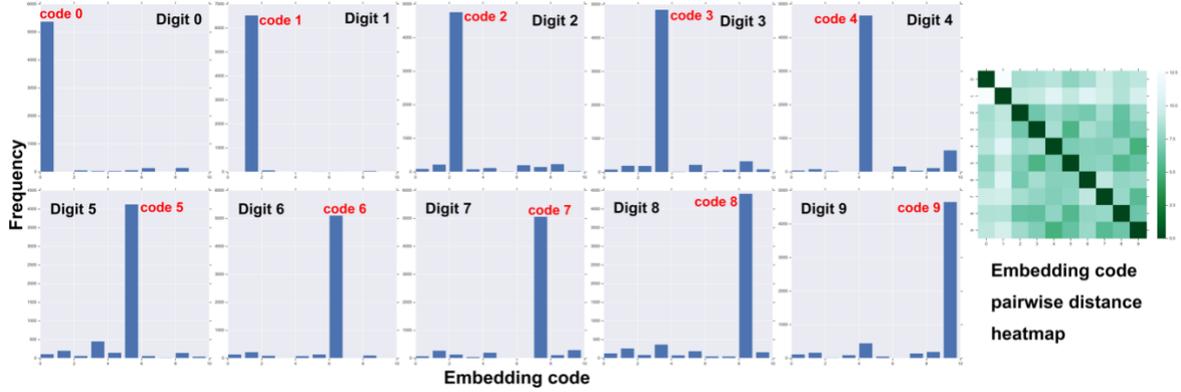

Fig. 3. Nearest embedding code for MNIST data and the pairwise distance between embedding codes. Left, the frequencies of nearest codes of images of each digit, visualized as bar plots. Right, the pairwise Euclidean distances between embedding codes visualized as a heatmap.

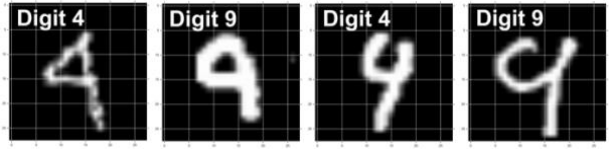

Fig. 4. Example MNIST images that show similarity between different digits.

## 5.2. *Application on the LINCS data*

We then applied S-VQ-VAE on the gene expression data from the LINCS project for learning associations between PCLs. The gene expression profile, as the outcome of communications between cellular functional modules through signaling networks, represents a more real-world data type compared to benchmark dataset like MNIST, thus is harder to learn. From Table 1, we can see that the standard VQ-VAE achieved a higher perplexity and a lower training error, while the S-VQ-VAE achieved a lower test error. Since the distribution of embedding codes is not predetermined in VQ-VAE, when the number of codes is big enough, the codes are given more flexibilities to accommodate the latent vector distribution than S-VQ-VAE; this explains its lower training error on the LINCS data. On the other hand, the lower test error of S-VQ-VAE indicates that the S-VQ-VAE model can be better generalized to unseen data given a limit amount of diverse training data, which is critical for learning representations for data from intractable distribution like gene expression profiles. Figure 5 gives the distribution of the nearest embedding code from VQ-VAE and S-VQ-VAE for samples of the same example PCLs. We can see that for S-VQ-VAE, most samples from the same PCL were correctly mapped to the code that was designated to learn a representation for this PCL. On the other hand, there is no such mapping enrichment in the VQ-VAE embedding codes, and it could be hard to interpret what each code may represent from a biological aspect.

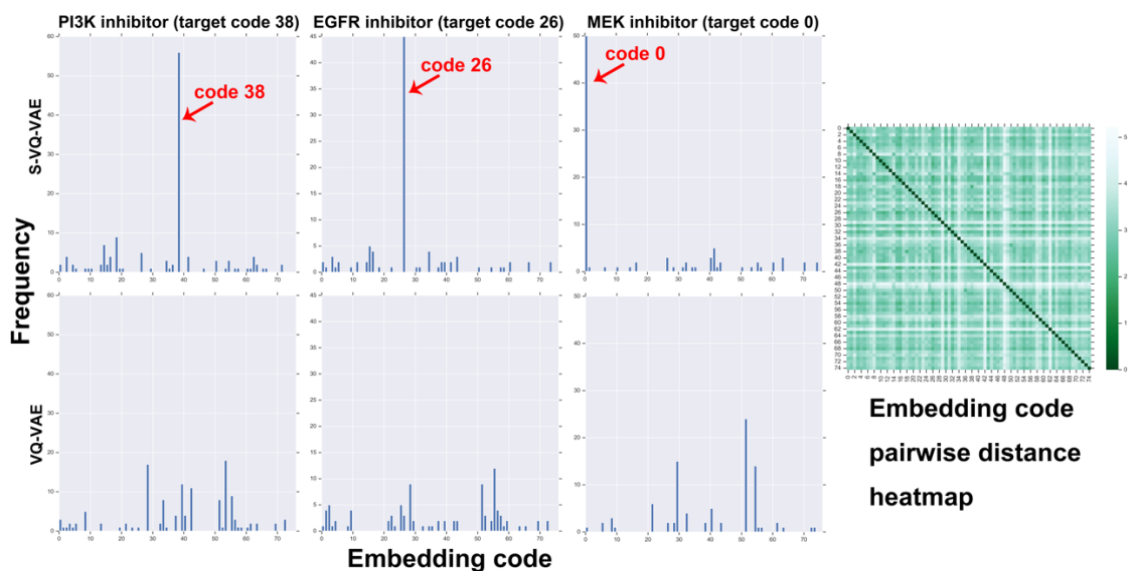

Fig. 5. Nearest embedding code for LINCS data and the pairwise distance between embedding codes. Left, the frequencies of nearest codes of samples from one of the three PCLs, PI3K inhibitor, EGFR inhibitor, and MEK inhibitor, visualized as bar plots. Right, the pairwise Euclidean distances between embedding codes visualized as a heatmap.

We next examined whether the representation learned for each PCL provides information about how these PCLs are correlated. From the distance heatmap in Figure 5, EGFR inhibitors and PDGFR-KIT inhibitors have the shortest pairwise distance, 2.01, followed by Raf inhibitors and PKC inhibitors with a distance of 2.05. The protein products of the *EGFR*, *PDGFR*, and *KIT* genes are known to affect a common set of downstream cell-growth-regulation signaling pathways, including the PI3K/AKT pathway and Ras pathway [17], and overexpression of *EGFR*, *PDGFR*, and *KIT* have been found in different types of cancer [18-22]. On the other hand, *PKC* has been known to directly regulate the activation of Raf-1 in cell signaling [23]. Therefore, it is not surprising that inhibiting the expression of these groups of genes would result in expression profiles sharing similar global characteristics. Figure 6 visualizes the nearest partner of each PCL based on the distance between corresponding embedding codes as a graph. Different colors mark different communities detected using the algorithm introduced in [24] (modularity score 0.742). The hubs of communities, including the EGFR inhibitors, Raf inhibitors, PKC inhibitors, and proteasome inhibitors, are known to affect core functional components in tumors [25-28]. Bidirectional correlations like the ones between EGFR inhibitors and PDGFR-KIT inhibitors were also found between several other pairs of PCLs, including PI3K inhibitors and mTOR inhibitors, and topoisomerase inhibitors and Aurora kinase inhibitors. PI3K inhibitors and mTOR inhibitors are known to impact the same cellular signaling pathway, the PI3K/AKT pathway [29]. Specifically, the mTOR, designating the mammalian target of rapamycin (a serine/threonine kinase), is a downstream effector of the PI3K/AKT pathway. Topoisomerase II has been found to regulate the activity of Aurora B kinase [30, 31]. All these direct functional interactions support that S-VQ-VAE is able to learn meaningful global representations from gene expression profiles of small molecule perturbed cells that reveal associations between mechanisms of action of different PCLs.

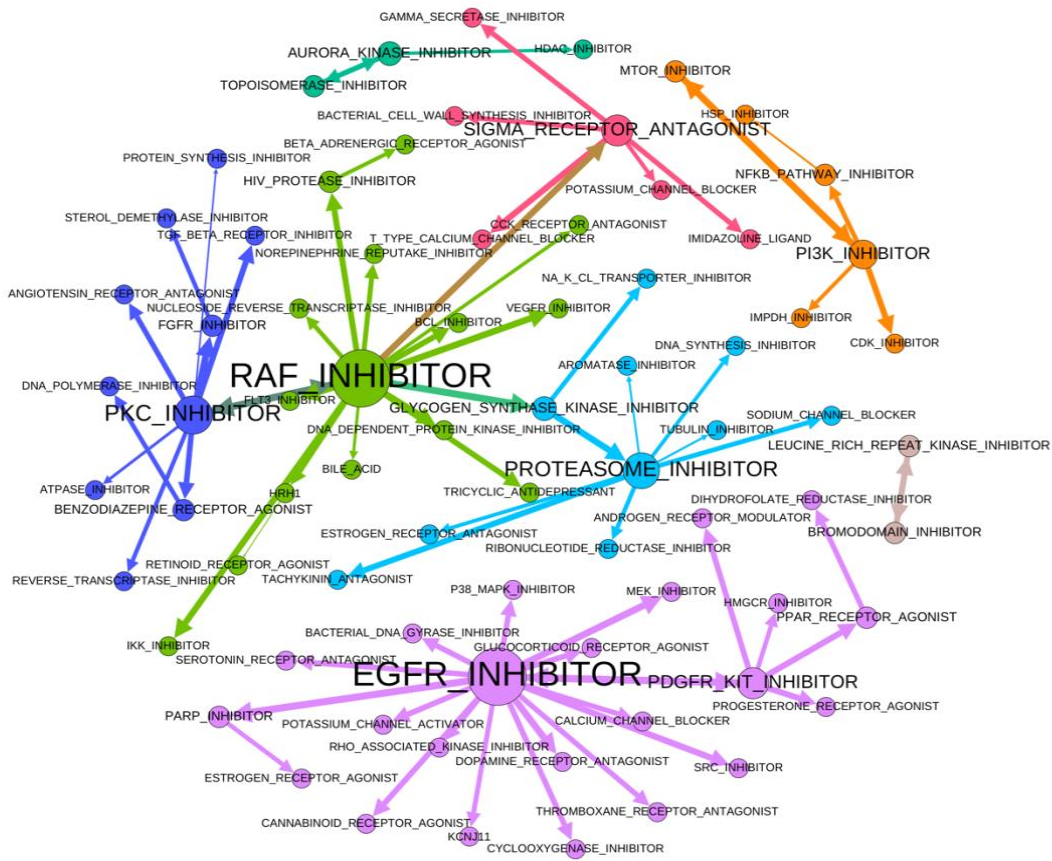

Fig. 6. Correlations between PCLs visualized as a graph. A directed edge in the graph indicates that the source node is the nearest node to the target code based on Euclidean distance. The node size is proportional to the out-degree. The edge width is inversely proportional to the distance. The color of a node indicates the modularity the node belongs to.

## 6. Discussion

Deep generative models are generally unsupervised models used to learn the joint distribution of the input data. To incorporate the individual latent representation learning and global characteristics learning into the same framework, we designed the S-VQ-VAE, which supplements VQ-VAE with a supervised learning component to learn a unique, interpretable global representation for each class of data. We have shown that the learned global representations can be used to reveal associations of different classes by applying the model on the MNIST and LINCS data. By training the embedding codes in a supervised manner, the number of codes can be readily determined based on the input data rather than chosen arbitrarily like in VQ-VAE, and the meaning of each node is predefined rather than interpreted after training. Given a limited number of embedding codes, S-VQ-VAE is more efficient than a standard VQ-VAE in representing the latent variable space and can be better generalized to unseen data. These properties would be beneficial to many application areas including drug development. Particularly, the associations between mechanisms of action of drugs discovered by S-VQ-VAE from the LINCS data can be used to guide new drug proposing and drug re-purposing for complex generic diseases like cancer. In addition, it is straightforward to extend S-VQ-VAE to semi-supervised scenarios by only updating the embedding space for labeled data while updating the encoder and decoder for both labeled and unlabeled data. The model can also be extended to multi-labeling data scenarios by allowing mapping to multiple embedding codes, and/or by adding an attention mechanism. We will explore these two directions in future work.